%% file: main.tex
\pdfoutput=1

\documentclass[11pt]{article}

\usepackage[]{ACL2023}

\usepackage{times} 
\usepackage{latexsym}

\usepackage[T1]{fontenc}

\usepackage[utf8]{inputenc}

\usepackage{microtype}

\usepackage{inconsolata}
\usepackage{graphicx}

\usepackage{microtype}
\usepackage{amsmath,amsfonts}

\usepackage{comment}
\usepackage{array}

\newcommand{\ie}{\emph{i.e., }}
\newcommand{\eg}{\emph{e.g., }}

\title{Multi-source Semantic Graph-based Multimodal Sarcasm Explanation Generation}

\author{Liqiang Jing$^1$, {\bf Xuemeng Song}$^1$\thanks{~~Xuemeng Song is the corresponding author.}, {\bf Kun Ouyang}$^1$, {\bf Mengzhao Jia}$^1$, {\bf Liqiang Nie}$^2$ \\   $^1$Shandong University\\ $^2$Harbin Institute of Technology (Shenzhen) \\ \{jingliqiang6, sxmustc, kunouyang10, jiamengzhao98, nieliqiang\}@gmail.com}

\begin{document}
\maketitle
\begin{abstract}
Multimodal Sarcasm Explanation (MuSE) is a new yet challenging task, which aims to generate a natural language sentence for a multimodal social post (an image as well as its caption) to explain why it contains sarcasm. 
Although the existing pioneer study has achieved great success with the BART backbone, it overlooks the gap between the visual feature space and the decoder semantic space, the object-level metadata of the image, as well as the potential external knowledge.
To solve these limitations, in this work, we propose a novel mulTi-source sEmantic grAph-based Multimodal sarcasm explanation scheme, named TEAM. In particular, TEAM extracts the object-level semantic meta-data instead of the traditional global visual features from the input image. Meanwhile, TEAM resorts to ConceptNet to obtain the external related knowledge concepts for the input text and the extracted object meta-data. 
Thereafter, TEAM introduces a multi-source semantic graph that comprehensively characterize the multi-source (\ie caption, object meta-data, external knowledge) semantic relations to facilitate the sarcasm reasoning. 
Extensive experiments on a public released dataset MORE verify the superiority of our model over cutting-edge methods.
\end{abstract}

\input{1_intro.tex}
\input{2_related_work}

\input{3_method}

\input{4_exp}

\input{5_conclusion}
\section*{Acknowledgements}
This work is supported by the Shandong Provincial Natural Science Foundation, No.:ZR2022YQ59; the National Natural Science Foundation of China, No.:62236003 and No.:62172261.

\section*{Limitations}
Our work mainly suffers from two key limitations. 1) Ignore that the text embedded in the image could also reflect the sarcastic intention.  As mentioned previously, we found that our model performs better on Non-OCR samples than the OCR samples. This may be due to the fact that our model ignores the text embedded in the image. Nevertheless, such embedded text could also indicate the ironic intention, (see Figure 3 (a)). We believe recognizing the text of the image can boost the performance of existing multimodal sarcasm explanation models. 2) Ignore that different knowledge concepts may contribute differently to the sarcasm reasoning. As shown in  Figure 3 (b), the related concepts ``disgusting'' and ``pleasant'' should contribute more than the concept ``night'' in the sarcasm reasoning. Currently, our model equally treats all the knowledge concepts. 


\bibliography{reference}
\bibliographystyle{acl_natbib}




\end{document}

%% file: 1_intro.tex
\section{Introduction}
Sarcasm is a common linguistic phenomenon, especially in posts on online social media platforms, that expresses people's emotions or opinions in a contrary manner. Since it benefits various real-world applications, such as customer feedback analysis and public opinion analysis, the sarcasm detection task has gained increasing research attention~\cite{DBLP:conf/acl/JoshiSB15, DBLP:conf/acl/AbercrombieH16}. Despite related great studies of the task, they can only identify the sarcastic post but could not give the concrete explanation for why it is sarcastic, making their detection results less convincing. 

\begin{figure}[t]
    \centering
    \includegraphics[width=0.95\linewidth]{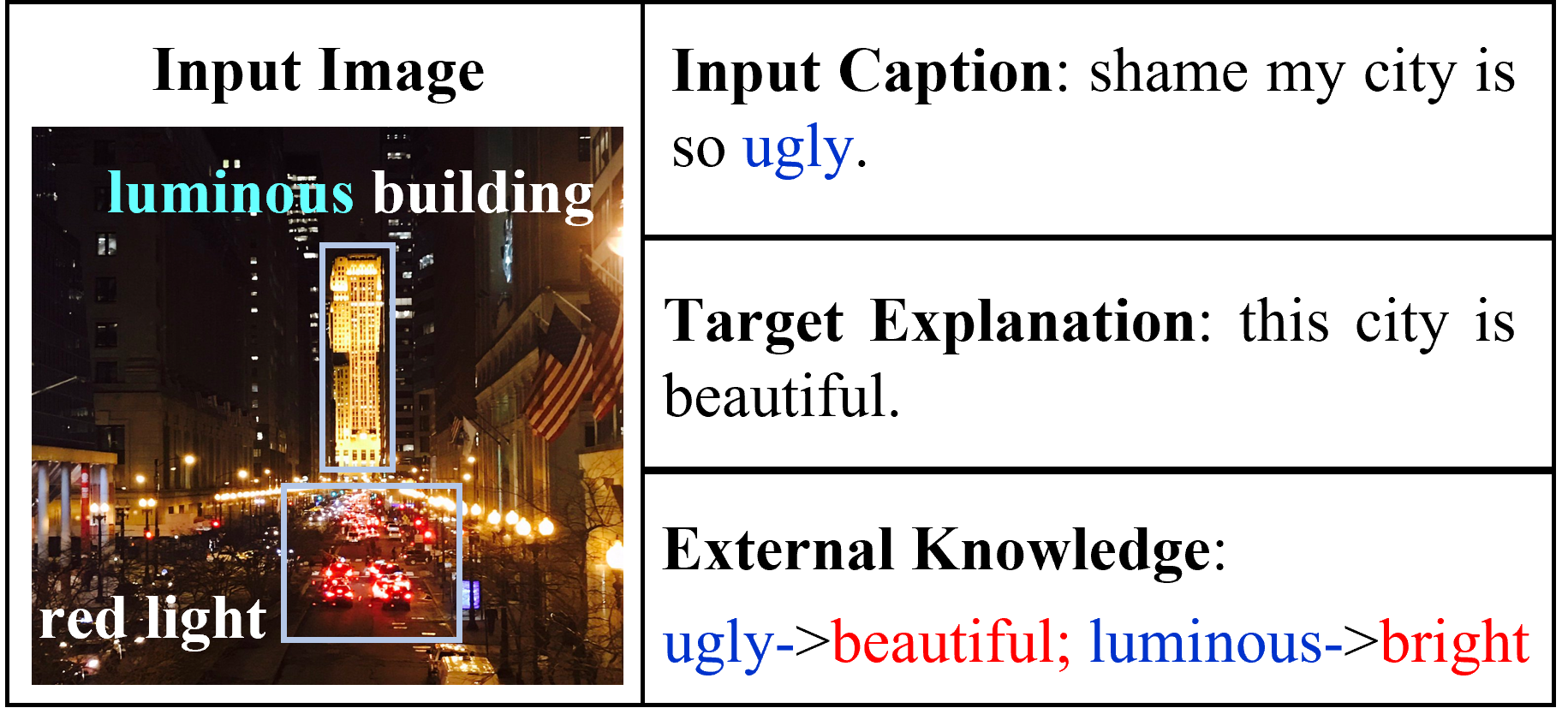}
    \caption{An example of the sarcasm explanation from MORE~\cite{DBLP:conf/aaai/Desai0A22}.  The key objects in the image are marked and the external knowledge is provided.}
    \vspace{-1em}
    \label{fig:fig2}
\end{figure}

Noticing this issue,  recent studies have shifted to the task of sarcasm explanation, which aims to generate a natural language sentence to explain the intended irony in a sarcastic post. 
For example, \citeauthor{DBLP:conf/acl/PeledR17} utilized the Recurrent Neural Network~(RNN)~\cite{DBLP:conf/sigdial/GhoshFM17}-based encoder-decoder architecture to tackle the sarcasm interpretation task. Although previous studies have attained impressive results, they focus on investigating the sarcasm explanation purely based on the textual input. Nevertheless, with the advances of multimedia devices, people tend to express their emotions or opinions through multimodal social posts. Moreover, the visual content usually also conveys important clues for explaining the sarcasm, as shown in Figure~\ref{fig:fig2}. Motivated by this, \citeauthor{DBLP:conf/aaai/Desai0A22} proposed the task of multimodal sarcasm explanation, which aims to generate the explanation for a multimodal input (\ie an image plus its corresponding caption). The authors gave a solution that first fuses the multimodal features with a cross-modal attention module, and then generates the explanation with the decoder of BART, a popular generative \mbox{pretrained} language model. 
Although this pioneer study has achieved promising performance, it still suffers from three key limitations.
\begin{itemize}
 \item  \textbf{L1: Overlook the gap between the visual feature space and the decoder semantic space.} The existing method directly adopts the visual feature of the input image with the context of BART decoder. In fact, the visual features may not match the semantic space of the BART well since it is pretrained only on the textual corpus, and these existing methods could not maximize the generation capacity of BART. 
 \item \textbf{L2: Overlook the object-level metadata of the image.} The existing work only extracts the global feature of the image, ignoring that only the key objects in the image relevant to the input caption contribute to sarcasm explanation (\eg ``luminous building'' and ``red light'' in Figure~\ref{fig:fig2}). Moreover, the object's metadata, \eg the class and attribute, which conveys important clues for the semantic understanding of the visual modality, merits our attention.
\item \textbf{L3: Overlook the potential external knowledge.} 
     The pioneer study fails to utilize the related knowledge contained in the external public knowledge base. 
     As shown in Figure~\ref{fig:fig2}, the related knowledge concepts obtained from ConceptNet~\cite{DBLP:conf/acl/GhosalHRMMP20} can strengthen the context learning (\eg bright)  and promote the explanation generation  (\eg beautiful). 
\end{itemize}

To tackle these limitations, we propose a novel mulTi-source sEmantic grAph-based Multimodal sarcasm explanation generation scheme, TEAM for short, which explores three semantic sources: the input caption, object meta-data derived from the input image, as well as the external knowledge. Specifically, TEAM includes four components: vision-based object-level semantic extraction, external related knowledge acquisition, multi-source semantic graph-based sarcasm reasoning, and sarcasm explanation generation. As shown in Figure~\ref{fig:framework}, in the first module, we focus on extracting the semantic meta-data of the key objects in the input image instead of the conventional global visual features, to adapt the decoding space of BART and facilitate the fine-grained sarcasm reasoning. In the second module, we target at acquiring the external related knowledge concepts for the input caption and the extracted object meta-data, where a large-scale knowledge base ConceptNet~\cite{DBLP:conf/acl/GhosalHRMMP20} is used as the reference. 
In the third module,  we construct the multi-source semantic graph to model the various semantic relations residing in the three semantic sources, and adopt GCN to fulfil the sarcasm reasoning. In the last module, we generate the target sarcasm explanation with the BART~\cite{DBLP:conf/acl/LewisLGGMLSZ20} decoder based on the three semantic sources.
We conduct extensive experiments on a public released multimodal sarcasm explanation dataset, on which our method outperforms the best baseline by 28.90 and 22.47 in terms of BLEU-4~\cite{DBLP:conf/acl/PapineniRWZ02} and ROUGE-L~\cite{lin-2004-rouge}, respectively.

Our contributions can be concluded as follows.
\begin{itemize}
    \item We propose a novel mulTi-source sEmantic grAph-based Multimodal sarcasm explanation scheme, where the fine-grained semantic information of the visual modality and the external knowledge concepts are jointly incorporated. 
    \item As far as we know, we are the first to adopt the object-level metadata of the visual modality to promote the multimodal sarcasm explanation generation by the generative pre-trained language model.
    \item We propose a multi-source semantic graph, which is able to comprehensively capture the semantic relation among the input caption, input image, and external knowledge concepts. As a byproduct, we release our code and parameters\footnote{\url{https://github.com/LiqiangJing/TEAM}.} to facilitate this community.
\end{itemize}

%% file: 2_related_work.tex
\begin{figure*}
    \centering
    \includegraphics[width=\textwidth]{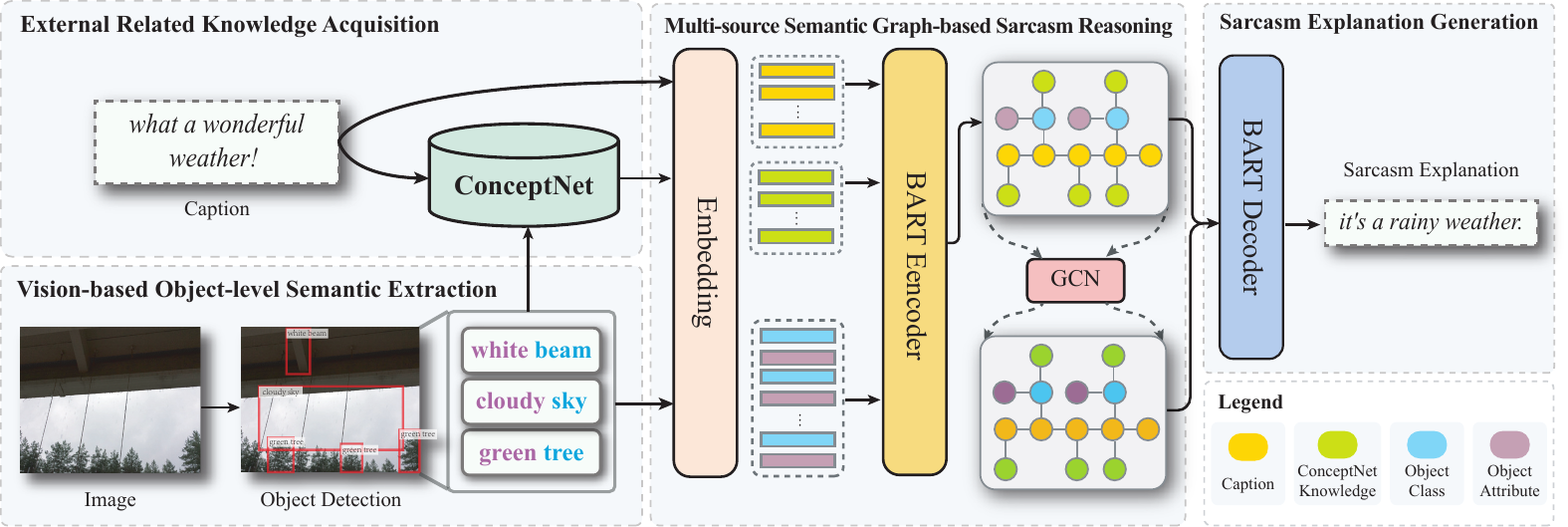}
    \caption{The architecture of the proposed TEAM, which consists of four key components: Vision-based Object-level Semantic Extraction, External Related Knowledge Acquisition, Multi-source Semantic Graph-based Sarcasm Reasoning, and Sarcasm Explanation Generation.}
    \label{fig:framework}
\end{figure*}

\section{Related Work}
Our work is related to sarcasm detection and sarcasm-related generation.
\subsection{Sarcasm Detection}
Sarcasm detection aims to detect whether a post contains the sarcasm meaning. 
Early studies on sarcasm detection~\cite{DBLP:journals/access/BouaziziO16,DBLP:conf/emnlp/FelboMSRL17} mainly use hand-crafted features, such as punctuation marks, POS tags, emojis, and lexicons, to detect the sarcastic intention. Later, with the development of deep learning techniques, some researchers resorted to neural network architectures for sarcasm detection~\cite{DBLP:conf/acl/SuTHL18,DBLP:conf/coling/BabanejadDAP20}.
Although these efforts have achieved promising progress, they focused on the text-based sarcasm detection, overlooking that the multimodal posts have been popping up all over the internet. 
Therefore, 
\citeauthor{DBLP:conf/mm/SchifanellaJTC16}~firstly proposed the \mbox{multimodal} sarcasm detection task and introduced a framework that fuses the textual and visual information with Convolutional Neural Networks~\cite{DBLP:conf/iccv/MaLSL15} to detect the sarcasm intention. One limitation of this work is that it ignored the fine-grained ironic semantic relation in the multimodal input.
Consequently, to boost the model performance, the following research efforts~\cite{DBLP:conf/aaai/qiao23,DBLP:conf/acl/KumarKA022,DBLP:conf/acl/ChakrabartyGMP20} resort to the Graph Convolutional Networks~(GCNs)~\cite{DBLP:conf/iclr/KipfW17} to mine inter-modal and intra-modal semantic association. 
Nevertheless, these efforts can only recognize whether a multimodal post contains the sarcastic meaning, but cannot explain why it is sarcastic, which is also important for various applications~\cite{DBLP:conf/aaai/Desai0A22}.

\subsection{Sarcasm-related Generation}
Apart from sarcasm detection, a few efforts attempted to conduct the sarcasm analysis by generating natural language. For example, some studies~\cite{DBLP:conf/acl/PeledR17,DBLP:conf/comad/DubeyJB19} resorted to machine translation models to generate non-sarcastic interpretation for the sarcastic text, which can help the smart customer service understand users' sarcastic comments and posts  on various platforms. In addition, \citeauthor{DBLP:conf/emnlp/MishraTS19} employed unsupervised methods to transform a negative sentiment sentence to a sarcastic text in the context of dialog systems, which can make the agent's responses more natural and attractive to the user.
Notably, these methods also only focus on text-based generation. 
Beyond them, recently, \citeauthor{DBLP:conf/aaai/Desai0A22} first proposed the multimodal sarcasm explanation task to support the sarcasm analysis and released a dataset, whose explanations are manually annotated. 
This method adopts the generative language model BART as the backbone, where the the global visual feature of the input image is incorporated with a cross-modal attention mechanism.
Despite its remarkable performance, this method overlooks the gap between the visual feature space and the BART decoder semantic space, the object-level metadata of the image, and the potential external knowledge, which are the major concerns of our model.

%% file: 3_method.tex

\section{Task Formulation}
Suppose we have a training dataset $\mathcal{D}$ composed of $N$ samples, \ie $\mathcal{D}=\{d_1,d_2,\cdots,d_N\}$. Each sample $d_i=\{T_i,V_i,Y_i\}$, where $T_i=\{t_1^i, t_2^i, \cdots t_{N_{t_i}}^i\}$ denotes the input caption which contains $N_{t_i}$ tokens, $V_i$ is the input image, and $Y_i=\{y_1^i, y_2^i, \cdots y_{N_{y_i}}^i\}$ denotes the target explanation text consisting of $N_{y_i}$ tokens. Notably, $N_{t_i}$ and $N_{y_i}$ vary on different samples. Based on these training samples, our target is to learn a multimodal sarcasm explanation model $\mathcal{F}$ that is able to generate the sarcasm explanation based on the given multimodal input as follows,
\begin{equation}
    \hat{Y_i} = \mathcal{F}(T_i,V_i|\Theta)
\end{equation}
where $\Theta$ is a set of to-be-learned parameters of the model $\mathcal{F}$. $\hat{Y_i}$ is the generated explanation text by 
 $\mathcal{F}$. For simplicity, we temporally omit the subscript $i$ that indexes the training samples.

\section{Method}
In this section, we detail the 
 four components of the proposed TEAM, as shown in Figure~\ref{fig:framework}. 

\subsection{Vision-based Object-level Semantic Extraction}
Considering that only the key visual information (\ie the objects in images) can demonstrate the sarcasm semantic, we propose to extract the object-level features of the image. Specifically, we feed the image into the Faster-RCNN \cite{DBLP:conf/cvpr/00010BT0GZ18}. Then for each region, it outputs not only the visual features (\eg content feature and positional feature) but also certain textual labels (\eg object class and object attribute). In our context, we only adopt the textual output, since we believe that textual labels contain rich semantics regarding the object, which should be beneficial towards the sarcasm reasoning, and fit better with the following encoding of the BART. Moreover, to ensure the quality of extracted object-level semantics, we only keep the top $K$ regions with the highest confidence.
Accordingly, for each image, we can obtain $K$ objects, each of which is associated with a class name and an attribute value. Formally, we have,
\begin{equation}
    \{(o_1,a_1),\cdots,(o_{K},a_{K})\} = \operatorname{F-RCNN}(V)
\end{equation}
where $o_j$ and $a_j$ are the extracted object class and attribute of the $j$-th object, respectively. 

\subsection{External Related Knowledge Acquisition}
As aforementioned, the knowledge inferred by the input caption can support the sarcasm explanation generation since it may supply some concepts that appeared in the explanation or help the ironic semantic understanding with some sentiment knowledge. 
Specifically, we choose ConceptNet that describes general human knowledge in graph format\footnote{\url{https://conceptnet.io/}.} as the source of external knowledge, 
which involves $3.1$ million concepts, and $38$ million relations. Given our context of sarcasm explanation generation, we 
adopt the preprocessed ConceptNet~\cite{DBLP:conf/aaai/LiLRRC22} that particularly covers the commonsense knowledge and emotional lexical knowledge, which plays an important role in the sarcasm reasoning.

To acquire the related external knowledge for the given multimodal input, \ie $(T, V)$, we first identify all the concepts in ConceptNet that are mentioned in the input caption and the object meta-data (\ie object class and object attribute) derived by Faster-RCNN. Let $\{c_1, \cdots, c_{N_c}\}$ be the set of identified concepts, where  $N_c$ is the total number of  identified concepts. 
We then use these identified concepts as the anchors to obtain the related concepts as the external knowledge for the multimodal input. Specifically, for each anchor concept $e$, we retrieve all its one-hop neighboring concepts from the knowledge graph ConceptNet and deem them as the external knowledge for $c$. 
Mathematically, let $\mathcal{N}(c)$ be the set of neighboring concepts of the concept $c$ in ConceptNet.  Then the related external knowledge for the multimodal input can be represented as $\{\mathcal{N}_{c_1},\mathcal{N}_{c_2}, \cdots, \mathcal{N}_{c_{N_c}}\}$.

\subsection{Multi-source Semantic Graph-based Sarcasm Reasoning}
 By now, we have three kinds of semantic sources: original input caption, object textual meta-data extracted from the input image, and external related textual concepts.  
To extract their features, we resort to the BART encoder, which has achieved compelling success on various natural language processing tasks, such as sentiment analysis~\cite{DBLP:conf/wanlp/MahdaouyMEMBK21} and multimodal summarization~\cite{DBLP:conf/acl/XingSMLMW20}. Since the three semantic sources share the same token form, we first concatenate them into a sequence of tokens, denoted as  $X$, and then feed $X$ into the BART encoder $\mathcal{E}$ as follows,
 
\begin{equation}
   \mathbf{H}=\mathcal{E}(X),\label{eq5}
\end{equation}
where $\mathbf{H} \in \mathbb{R}^{N \times D}$ is the encoded representation matrix, each column of which corresponds to a token, and $N$ is the total number of tokens in $X$.

In fact, there are rich semantic relations resided in the three kinds of semantic sources that can be used for the sarcasm reasoning and the corresponding explanation generation.  For example, the semantic correlation among tokens in the input caption can help the intra-modal inconsistency mining; 
the semantic correspondence between tokens in the input caption and that in the object meta-data can facilitate the cross-modal inconsistency uncovering. Moreover, linking the retrieved knowledge concepts to tokens in the input caption as well as those in the object meta-data promotes the semantic understanding of the multimodal input. 

In light of this, for each sample $d$, we propose to construct a 
 multi-source semantic graph $\mathcal{G}$ to comprehensively capture the above semantic relations. Let $\mathcal{H}=\{{h}_1,\cdots, {h}_N\}$ denote the set of nodes, which correspond to $N$ tokens in $X$ and can be divided into three categories: textual caption nodes, 
 object nodes, and knowledge nodes. The representations of these nodes are initialized by $\mathbf{H}$.

\begin{figure}
    \centering
    \includegraphics[width=0.5\textwidth]{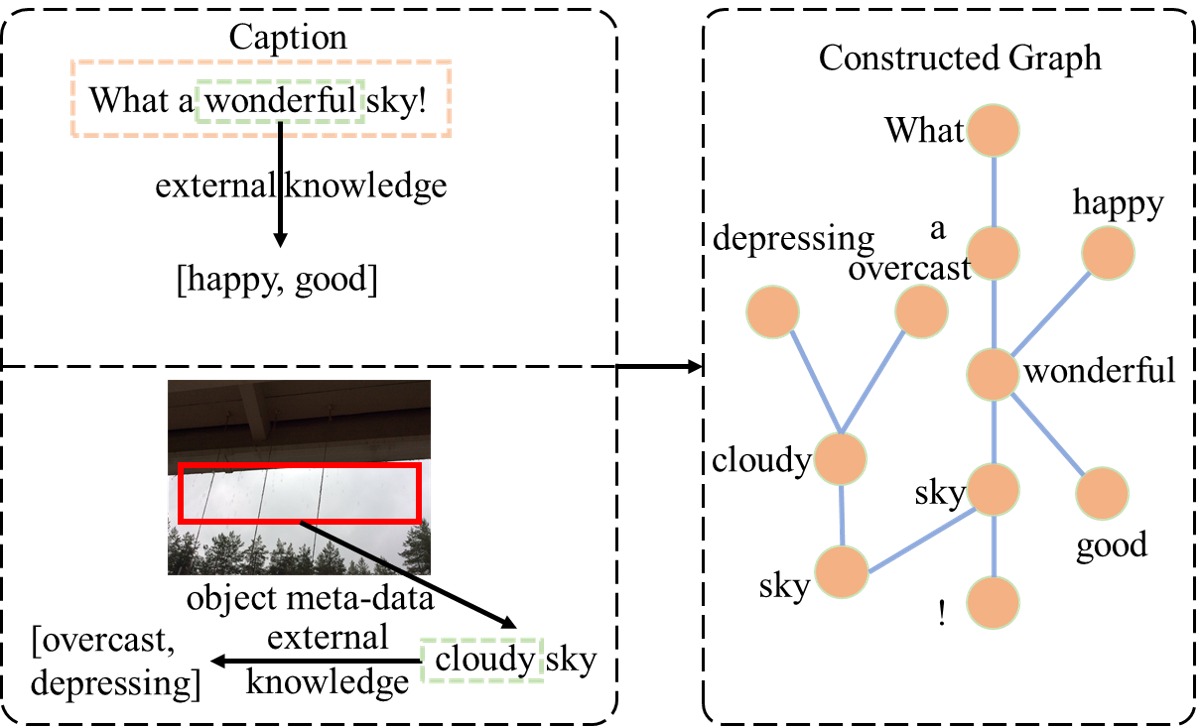}
    \caption{An example of the multi-source semantic graph construction process.}
    \label{fig:sample}
\end{figure}

The edges of this graph are defined according to the semantic relations among these nodes as follows. 1) We first link the semantically correlated text nodes by adding an edge between each pair of adjacent tokens in the input caption. 
2) We then introduce an edge between each object class and its corresponding object attribute, to link the object nodes that characterize the same object. 3) To capture the cross-modal semantic relation, we build an edge between each object class and its most similar token in the input caption, where the cosine similarity metric is used. And 4) for each retrieved knowledge concept, we link it with tokens in the input caption and object meta-data that act as the anchor concept in the aforementioned knowledge concept retrieval process. Formally, let $\mathbf{A}\in \mathbb{R}^{N\times N}$ denote the adjacency matrix of our constructed multi-source semantic graph. In order to facilitate understanding, we describe the construction process of the multi-source semantic graph in Figure~\ref{fig:sample}. 

Thereafter, we resort to the commonly used GCNs to conduct the sarcasm reasoning. 
Specifically, suppose we adopt $L$ layers of GCN. Then all the node representations are iteratively updated as follows,
\begin{equation}
    \mathbf{G}_{l}=ReLU(\tilde{\mathbf{A}}\mathbf{G}_{l-1}\mathbf{W}_l), l \in [1,L],
\end{equation}
where $\tilde{\mathbf{A}} = (\mathbf{D})^{-\frac{1}{2}}\mathbf{A}(\mathbf{D})^{-\frac{1}{2}}$ is the normalized symmetric adjacency matrix, and $\mathbf{D}$ is the degree matrix of $\mathbf{A}$. In addition, $\mathbf{W}_l \in \mathbb{R}^{D \times D}$ is a trainable parameter of the $l$-th GCN layer.
$\mathbf{G}_l$ are the representations of nodes obtained in the $l$-th layer GCN, where 
$\mathbf{G}_0=\mathbf{H}$ is the initial node representation. 

\subsection{Sarcasm Explanation Generation}

The final nodes representation $\mathbf{G}_L$ obtained by the $L$-layer GCN should absorb rich semantic information from their correlated nodes and can be used as the input for the following sarcasm explanation generation. Considering that the residual connection always performs well in the task of 
text generation~\cite{DBLP:conf/nips/VaswaniSPUJGKP17}, we also introduce a residual connection for generating the sarcasm explanation. Specifically, we first fuse the initial and final nodes representations as follows,
\begin{equation}
\mathbf{R}=\mathbf{H}+\mathbf{G}_L
\end{equation}
where $\mathbf{R} \in \mathbb{R}^{N\times D}$ denotes the fused node representation.
We then feed $\mathbf{R}$ 
to the decoder of the pre-trained BART. The decoder works in an auto-regressive manner, namely, producing the next word by considering all the previously decoded outputs  as follows,
\begin{equation}
\hat{\textbf{y}}_t=BART\_Decoder(\mathbf{R}, \hat{Y}_{<t}),
\end{equation}
where $t \in [1,N_y]$ and $\hat{\textbf{y}}_t \in \mathbb{R}^{|\mathcal{V}|}$ is the predicted $t$-th token's probability distribution of the target sarcasm explanation. $\hat{Y}_{<t}$ refers to the previously predicted $t$-$1$ tokens. Notably, in the training phase, to avoid the accumulated error, $\hat{Y}_{<t}$ will be replaced by ${Y}_{<t}$, \ie the previous $t-1$ tokens in the target sarcasm explanation. 

For optimizing our TEAM, we adopt the standard cross-entropy loss function as follows,
\begin{equation} \label{prob}
  \mathcal{L}_{Gen} = -1/N_y\sum_{i=1}^{N_y}\log (\hat{\textbf{y}}_i[t]),
 \end{equation}
where $\hat{\textbf{y}}_i[t]$ is the element of  $\hat{\textbf{y}}_i$ that corresponds to the $i$-th token of the target explanation, and $N_y$ is the total number of tokens in the target sarcasm explanation $Y$. 

%% file: 4_exp.tex
\begin{table}[tb]
\caption{Statistics of the MORE dataset. Avg.length and |$\mathcal{V}$| denote the average length of text and the vocabulary size, respectively.}
\label{tab:table1}
\resizebox{\columnwidth}{!}{%
\begin{tabular}{l|c|cc|cc}
\hline
{\color[HTML]{333333} }                                 &                                        & \multicolumn{2}{c|}{Caption}            & \multicolumn{2}{c}{Explanation}        \\ \cline{3-6} 
    {\multirow{-2}{*}{\textbf{Name}}}& \multirow{-2}{*}{\textbf{\#Samples}} & \multicolumn{1}{c|}{Avg.length} & |$\mathcal{V}$|   & \multicolumn{1}{c|}{Avg.length} & |$\mathcal{V}$| \\ \hline
Train                                                   & 2,983                                   & \multicolumn{1}{c|}{19.75}      & 9,677  & \multicolumn{1}{c|}{15.47}      & 5,972 \\
Val                                                     & 175                                    & \multicolumn{1}{c|}{18.85}      & 1,230  & \multicolumn{1}{c|}{15.39}      & 922  \\
Test                                                    & 352                                    & \multicolumn{1}{c|}{19.43}      & 2,172  & \multicolumn{1}{c|}{15.08}      & 1,527 \\ \hline
Total                                                   & 3,510                                   & \multicolumn{1}{c|}{19.68}      & 10,865 & \multicolumn{1}{c|}{15.43}      & 6,669 \\ \hline
\end{tabular}
}
\end{table}

\begin{table*}[tb]
\centering
\footnotesize
\caption{Comparing the generation performance of our model against state-of-the-art baselines on the MORE dataset. The best results are in boldface, while the second best are underlined.}
\label{tab:table2}
\resizebox{\textwidth}{!}{
\begin{tabular}{lcccccccccccc}
\hline
\multicolumn{1}{l|}{{\color[HTML]{333333} }}                                 & \multicolumn{4}{c|}{\textbf{BLEU}}                                                                                                                                        & \multicolumn{3}{c|}{\textbf{Rouge}}                                                                             & \multicolumn{1}{c|}{}                                  & \multicolumn{3}{c|}{\textbf{BERT-Score}}                                                                     & \textbf{Sent-BERT} \\ \cline{2-8} \cline{10-13} 
\multicolumn{1}{l|}{\multirow{-2}{*}{{\color[HTML]{333333} \textbf{Model}}}} & \multicolumn{1}{c|}{B1}                                 & \multicolumn{1}{c|}{B2}             & \multicolumn{1}{c|}{B3}             & \multicolumn{1}{c|}{B4}             & \multicolumn{1}{c|}{RL}             & \multicolumn{1}{c|}{R1}             & \multicolumn{1}{c|}{R2}             & \multicolumn{1}{c|}{\multirow{-2}{*}{\textbf{METEOR}}} & \multicolumn{1}{c|}{Pre}           & \multicolumn{1}{c|}{Rec}           & \multicolumn{1}{c|}{F1}            & (Cosine)           \\ \hline
\multicolumn{1}{l|}{PGN}                               & \multicolumn{1}{c|}{17.54}                              & \multicolumn{1}{c|}{6.31}           & \multicolumn{1}{c|}{2.33}           & \multicolumn{1}{c|}{1.67}           & \multicolumn{1}{c|}{16.00}          & \multicolumn{1}{c|}{17.35}          & \multicolumn{1}{c|}{6.90}           & \multicolumn{1}{c|}{15.06}                             & \multicolumn{1}{c|}{84.80}          & \multicolumn{1}{c|}{85.10}          & \multicolumn{1}{c|}{84.90}          & 49.42              \\
\multicolumn{1}{l|}{Transformer}                                             & \multicolumn{1}{c|}{11.44}                              & \multicolumn{1}{c|}{4.79}           & \multicolumn{1}{c|}{1.68}           & \multicolumn{1}{c|}{0.73}           & \multicolumn{1}{c|}{15.90}          & \multicolumn{1}{c|}{17.78}          & \multicolumn{1}{c|}{5.83}           & \multicolumn{1}{c|}{9.74}                              & \multicolumn{1}{c|}{83.40}          & \multicolumn{1}{c|}{84.90}          & \multicolumn{1}{c|}{84.10}          & 52.55              \\ \hline
\multicolumn{1}{l|}{MFFG-RNN}                                                & \multicolumn{1}{c|}{14.16}                              & \multicolumn{1}{c|}{6.10}           & \multicolumn{1}{c|}{2.31}           & \multicolumn{1}{c|}{1.12}           & \multicolumn{1}{c|}{16.21}          & \multicolumn{1}{c|}{17.47}          & \multicolumn{1}{c|}{5.53}           & \multicolumn{1}{c|}{12.31}                             & \multicolumn{1}{c|}{81.50}          & \multicolumn{1}{c|}{84.00}            & \multicolumn{1}{c|}{82.70}          & 44.65              \\
\multicolumn{1}{l|}{MFFG-Transf}                                             & \multicolumn{1}{c|}{13.55}                              & \multicolumn{1}{c|}{4.95}           & \multicolumn{1}{c|}{2.00}           & \multicolumn{1}{c|}{0.76}           & \multicolumn{1}{c|}{15.14}          & \multicolumn{1}{c|}{16.84}          & \multicolumn{1}{c|}{4.30}           & \multicolumn{1}{c|}{10.97}                             & \multicolumn{1}{c|}{81.10}          & \multicolumn{1}{c|}{83.80}          & \multicolumn{1}{c|}{82.40}          & 41.58              \\
\multicolumn{1}{l|}{M-Transf}                                                & \multicolumn{1}{c|}{14.37}                              & \multicolumn{1}{c|}{6.48}           & \multicolumn{1}{c|}{2.94}           & \multicolumn{1}{c|}{1.57}           & \multicolumn{1}{c|}{18.77}          & \multicolumn{1}{c|}{20.99}          & \multicolumn{1}{c|}{6.98}           & \multicolumn{1}{c|}{12.84}                             & \multicolumn{1}{c|}{86.30}          & \multicolumn{1}{c|}{86.20}          & \multicolumn{1}{c|}{86.20}          & 53.85              \\
\multicolumn{1}{l|}{ExMore}                                                  & \multicolumn{1}{c|}{19.26}                             & \multicolumn{1}{c|}{11.21}        & \multicolumn{1}{c|}{6.56}       & \multicolumn{1}{c|}{4.26}         & \multicolumn{1}{c|}{25.23}         & \multicolumn{1}{c|}{27.55}         & \multicolumn{1}{c|}{12.49}         & \multicolumn{1}{c|}{19.16}                            & \multicolumn{1}{c|}{88.30}       & \multicolumn{1}{c|}{87.50}         & \multicolumn{1}{c|}{87.90}         & 59.12             \\ \hline
\multicolumn{1}{l|}{TEAM-w/o-Know}                                                & \multicolumn{1}{c|}{\underline{52.63}}                              & \multicolumn{1}{c|}{\underline{42.42}}          & \multicolumn{1}{c|}{\underline{35.80}}          & \multicolumn{1}{c|}{\underline{30.91}}          & \multicolumn{1}{c|}{\underline{48.67}}          & \multicolumn{1}{c|}{\underline{49.28}}          & \multicolumn{1}{c|}{\underline{33.18}}          & \multicolumn{1}{c|}{\underline{48.53}}                             & \multicolumn{1}{c|}{\underline{90.90}}          & \multicolumn{1}{c|}{\underline{91.40}}          & \multicolumn{1}{c|}{\underline{91.10}}          & \underline{71.58}              \\

\multicolumn{1}{l|}{\textbf{TEAM}}                                      & \multicolumn{1}{c|}{\textbf{55.32}}                     & \multicolumn{1}{c|}{\textbf{45.12}} & \multicolumn{1}{c|}{\textbf{38.27}} & \multicolumn{1}{c|}{\textbf{33.16}} & \multicolumn{1}{c|}{\textbf{50.58}} & \multicolumn{1}{c|}{\textbf{51.72}} & \multicolumn{1}{c|}{\textbf{34.96}} & \multicolumn{1}{c|}{\textbf{50.95}}                    & \multicolumn{1}{c|}{\textbf{91.80}} & \multicolumn{1}{c|}{\textbf{91.60}} & \multicolumn{1}{c|}{\textbf{91.70}} & \textbf{72.92}     \\ \hline
\multicolumn{13}{c}{(a) All samples}                                                                                                           \\ \hline
\multicolumn{1}{l|}{{\color[HTML]{333333} }}                                 & \multicolumn{4}{c|}{\textbf{BLEU}}                                                                                                                                        & \multicolumn{3}{c|}{\textbf{Rouge}}                                                                             & \multicolumn{1}{c|}{}                                  & \multicolumn{3}{c|}{\textbf{BERT-Score}}                                                                     & \textbf{Sent-BERT} \\ \cline{2-8} \cline{10-13} 
\multicolumn{1}{l|}{\multirow{-2}{*}{{\color[HTML]{333333} \textbf{Model}}}} & \multicolumn{1}{c|}{B1}                                 & \multicolumn{1}{c|}{B2}             & \multicolumn{1}{c|}{B3}             & \multicolumn{1}{c|}{B4}             & \multicolumn{1}{c|}{RL}             & \multicolumn{1}{c|}{R1}             & \multicolumn{1}{c|}{R2}             & \multicolumn{1}{c|}{\multirow{-2}{*}{\textbf{METEOR}}} & \multicolumn{1}{c|}{Pre}           & \multicolumn{1}{c|}{Rec}           & \multicolumn{1}{c|}{F1}            & (Cosine)           \\ \hline
\multicolumn{1}{l|}{PGN}                               & \multicolumn{1}{c|}{17.87}                              & \multicolumn{1}{c|}{6.37}           & \multicolumn{1}{c|}{1.92}           & \multicolumn{1}{c|}{1.26}           & \multicolumn{1}{c|}{16.43}          & \multicolumn{1}{c|}{17.80}          & \multicolumn{1}{c|}{6.92}           & \multicolumn{1}{c|}{15.62}                             & \multicolumn{1}{c|}{84.70}          & \multicolumn{1}{c|}{85.20}          & \multicolumn{1}{c|}{84.90}          & 48.77              \\
\multicolumn{1}{l|}{Transformer}                                             & \multicolumn{1}{c|}{11.65}                              & \multicolumn{1}{c|}{5.65}           & \multicolumn{1}{c|}{1.73}           & \multicolumn{1}{c|}{0.69}           & \multicolumn{1}{c|}{16.16}          & \multicolumn{1}{c|}{17.41}          & \multicolumn{1}{c|}{6.26}           & \multicolumn{1}{c|}{10.13}                             & \multicolumn{1}{c|}{83.60}          & \multicolumn{1}{c|}{85.10}          & \multicolumn{1}{c|}{84.30}          & 48.40              \\ \hline
\multicolumn{1}{l|}{MFFG-RNN}                                                & \multicolumn{1}{c|}{15.43}                              & \multicolumn{1}{c|}{6.82}           & \multicolumn{1}{c|}{2.46}           & \multicolumn{1}{c|}{1.33}           & \multicolumn{1}{c|}{17.40}          & \multicolumn{1}{c|}{18.61}          & \multicolumn{1}{c|}{5.71}           & \multicolumn{1}{c|}{12.98}                             & \multicolumn{1}{c|}{81.60}          & \multicolumn{1}{c|}{84.30}          & \multicolumn{1}{c|}{82.90}          & 42.72              \\
\multicolumn{1}{l|}{MFFG-Transf}                                             & \multicolumn{1}{c|}{13.28}                              & \multicolumn{1}{c|}{5.35}           & \multicolumn{1}{c|}{1.49}           & \multicolumn{1}{c|}{0.26}           & \multicolumn{1}{c|}{14.90}          & \multicolumn{1}{c|}{16.80}          & \multicolumn{1}{c|}{4.35}           & \multicolumn{1}{c|}{11.19}                             & \multicolumn{1}{c|}{81.30}          & \multicolumn{1}{c|}{84.00}            & \multicolumn{1}{c|}{82.60}          & 41.68              \\
\multicolumn{1}{l|}{M-Transf}                                                & \multicolumn{1}{c|}{14.91}                              & \multicolumn{1}{c|}{6.90}           & \multicolumn{1}{c|}{2.66}           & \multicolumn{1}{c|}{0.83}           & \multicolumn{1}{c|}{19.34}          & \multicolumn{1}{c|}{21.05}          & \multicolumn{1}{c|}{7.08}           & \multicolumn{1}{c|}{13.91}                             & \multicolumn{1}{c|}{86.50}          & \multicolumn{1}{c|}{86.30}          & \multicolumn{1}{c|}{86.40}          & 51.77              \\ 
\multicolumn{1}{l|}{ExMore}                                                  & \multicolumn{1}{c|}{19.47}                             & \multicolumn{1}{c|}{11.69}         & \multicolumn{1}{c|}{6.82}        & \multicolumn{1}{c|}{4.27}          & \multicolumn{1}{c|}{24.92}         & \multicolumn{1}{c|}{27.12}         & \multicolumn{1}{c|}{12.12}         & \multicolumn{1}{c|}{19.20}                            & \multicolumn{1}{c|}{88.30}         & \multicolumn{1}{c|}{87.60}         & \multicolumn{1}{c|}{88.00}         & 56.95             \\ \hline
\multicolumn{1}{l|}{TEAM-w/o-Know}                                                & \multicolumn{1}{c|}{\underline{53.43}}                              & \multicolumn{1}{c|}{\underline{43.41}}          & \multicolumn{1}{c|}{\underline{36.77}}          & \multicolumn{1}{c|}{\underline{31.78}}          & \multicolumn{1}{c|}{\underline{49.72}}          & \multicolumn{1}{c|}{\underline{51.12}}          & \multicolumn{1}{c|}{\underline{34.78}}          & \multicolumn{1}{c|}{\underline{49.24}}                             & \multicolumn{1}{c|}{\underline{91.50}}          & \multicolumn{1}{c|}{\underline{91.90}}          & \multicolumn{1}{c|}{\underline{91.80}}          & \underline{71.62}              \\

\multicolumn{1}{l|}{\textbf{TEAM}}                                      & \multicolumn{1}{c|}{\textbf{56.45}}                     & \multicolumn{1}{c|}{\textbf{46.34}} & \multicolumn{1}{c|}{\textbf{39.58}} & \multicolumn{1}{c|}{\textbf{34.34}} & \multicolumn{1}{c|}{\textbf{52.79}} & \multicolumn{1}{c|}{\textbf{53.81}} & \multicolumn{1}{c|}{\textbf{36.78}} & \multicolumn{1}{c|}{\textbf{51.62}}                    & \multicolumn{1}{c|}{\textbf{92.40}} & \multicolumn{1}{c|}{\textbf{92.90}} & \multicolumn{1}{c|}{\textbf{92.30}} & \textbf{73.35}     \\ \hline
\multicolumn{13}{c}{(b) Non-OCR samples}                                                                                                    \\ \hline
\multicolumn{1}{l|}{{\color[HTML]{333333} }}                                 & \multicolumn{4}{c|}{\textbf{BLEU}}                                                                                                                                        & \multicolumn{3}{c|}{\textbf{Rouge}}                                                                             & \multicolumn{1}{c|}{}                                  & \multicolumn{3}{c|}{\textbf{BERT-Score}}                                                                     & \textbf{Sent-BERT} \\ \cline{2-8} \cline{10-13} 
\multicolumn{1}{l|}{\multirow{-2}{*}{{\color[HTML]{333333} \textbf{Model}}}} & \multicolumn{1}{c|}{{\color[HTML]{333333} \textbf{B1}}} & \multicolumn{1}{c|}{B2}             & \multicolumn{1}{c|}{B3}             & \multicolumn{1}{c|}{B4}             & \multicolumn{1}{c|}{RL}             & \multicolumn{1}{c|}{R1}             & \multicolumn{1}{c|}{R2}             & \multicolumn{1}{c|}{\multirow{-2}{*}{\textbf{METEOR}}} & \multicolumn{1}{c|}{Pre}           & \multicolumn{1}{c|}{Rec}           & \multicolumn{1}{c|}{F1}            & (Cosine)           \\ \hline
\multicolumn{1}{l|}{ PGN}                               & \multicolumn{1}{c|}{17.19}                              & \multicolumn{1}{c|}{6.08}           & \multicolumn{1}{c|}{2.49}           & \multicolumn{1}{c|}{1.79}           & \multicolumn{1}{c|}{15.55}          & \multicolumn{1}{c|}{16.92}          & \multicolumn{1}{c|}{6.76}           & \multicolumn{1}{c|}{14,64}                             & \multicolumn{1}{c|}{84.90}          & \multicolumn{1}{c|}{84.90}          & \multicolumn{1}{c|}{84.90}          & 49.53              \\
\multicolumn{1}{l|}{Transformer}                                             & \multicolumn{1}{c|}{10.68}                              & \multicolumn{1}{c|}{4.01}           & \multicolumn{1}{c|}{1.49}           & \multicolumn{1}{c|}{0.71}           & \multicolumn{1}{c|}{15.04}          & \multicolumn{1}{c|}{17.25}          & \multicolumn{1}{c|}{5.32}           & \multicolumn{1}{c|}{8.99}                              & \multicolumn{1}{c|}{83.20}          & \multicolumn{1}{c|}{84.70}          & \multicolumn{1}{c|}{83.90}          & 53.94              \\ \hline
\multicolumn{1}{l|}{MFFG-RNN}                                                & \multicolumn{1}{c|}{12.18}                              & \multicolumn{1}{c|}{4.92}           & \multicolumn{1}{c|}{1.73}           & \multicolumn{1}{c|}{0.88}           & \multicolumn{1}{c|}{14.01}          & \multicolumn{1}{c|}{15.18}          & \multicolumn{1}{c|}{4.56}           & \multicolumn{1}{c|}{10.64}                             & \multicolumn{1}{c|}{81.20}          & \multicolumn{1}{c|}{83.70}          & \multicolumn{1}{c|}{82.40}          & 45.91              \\
\multicolumn{1}{l|}{MFFG-Transf}                                             & \multicolumn{1}{c|}{12.87}                              & \multicolumn{1}{c|}{4.12}           & \multicolumn{1}{c|}{1.69}           & \multicolumn{1}{c|}{0.62}           & \multicolumn{1}{c|}{14.20}          & \multicolumn{1}{c|}{15.54}          & \multicolumn{1}{c|}{3.53}           & \multicolumn{1}{c|}{9.70}                              & \multicolumn{1}{c|}{81.00}            & \multicolumn{1}{c|}{83.60}          & \multicolumn{1}{c|}{82.30}          & 41.13              \\
\multicolumn{1}{l|}{M-Transf}                                                & \multicolumn{1}{c|}{14.06}                              & \multicolumn{1}{c|}{6.25}           & \multicolumn{1}{c|}{3.22}           & \multicolumn{1}{c|}{2.28}           & \multicolumn{1}{c|}{18.42}          & \multicolumn{1}{c|}{21.04}          & \multicolumn{1}{c|}{7.01}           & \multicolumn{1}{c|}{12.06}                             & \multicolumn{1}{c|}{86.20}          & \multicolumn{1}{c|}{86.10}          & \multicolumn{1}{c|}{86.10}          & 55.66              \\ 
\multicolumn{1}{l|}{ExMore}                                                & \multicolumn{1}{c|}{19.40}                              & \multicolumn{1}{c|}{11.31}           & \multicolumn{1}{c|}{6.83}           & \multicolumn{1}{c|}{4.76}           & \multicolumn{1}{c|}{25.66}          & \multicolumn{1}{c|}{28.02}          & \multicolumn{1}{c|}{{12.10}}           & \multicolumn{1}{c|}{19.15}                             & \multicolumn{1}{c|}{88.20}          & \multicolumn{1}{c|}{87.50}          & \multicolumn{1}{c|}{87.90}          & 60.82              \\ \hline
\multicolumn{1}{l|}{TEAM-w/o-Know}                                                & \multicolumn{1}{c|}{\underline{51.91}}                              & \multicolumn{1}{c|}{\underline{41.51}}          & \multicolumn{1}{c|}{\underline{34.85}}          & \multicolumn{1}{c|}{\underline{29.85}}          & \multicolumn{1}{c|}{\underline{47.53}}          & \multicolumn{1}{c|}{\underline{49.00}}          & \multicolumn{1}{c|}{\underline{32.77}}          & \multicolumn{1}{c|}{\underline{47.94}}                             & \multicolumn{1}{c|}{\underline{90.50}}          & \multicolumn{1}{c|}{\textbf{91.00}}          & \multicolumn{1}{c|}{\underline{90.70}}          & \underline{71.43}              \\
\multicolumn{1}{l|}{\textbf{TEAM}}                                      & \multicolumn{1}{c|}{\textbf{52.88}}                     & \multicolumn{1}{c|}{\textbf{43.08}} & \multicolumn{1}{c|}{\textbf{36.81}} & \multicolumn{1}{c|}{\textbf{32.34}} & \multicolumn{1}{c|}{\textbf{48.46}} & \multicolumn{1}{c|}{\textbf{49.68}} & \multicolumn{1}{c|}{\textbf{33.83}} & \multicolumn{1}{c|}{\textbf{49.25}}                    & \multicolumn{1}{c|}{\textbf{90.90}} & \multicolumn{1}{c|}{\underline{90.00}} & \multicolumn{1}{c|}{\textbf{90.80}} & \textbf{71.93}     \\ \hline
\multicolumn{13}{c}{(c) OCR samples}              
\end{tabular}%
}
\end{table*}

\section{Experiment}

\begin{table*}[tb]
\centering
\caption{Experiment results of ablation study. The best results are in boldface.}
\footnotesize
\label{tab:ablation}
\resizebox{\textwidth}{!}{%
\begin{tabular}{lcccccccccccc}
\hline
\multicolumn{1}{l|}{{\color[HTML]{333333} }}                                 & \multicolumn{4}{c|}{\textbf{BLEU}}                                                                                                                                        & \multicolumn{3}{c|}{\textbf{Rouge}}                                                                             & \multicolumn{1}{c|}{}                                  & \multicolumn{3}{c|}{\textbf{BERT-Score}}                                                                     & \textbf{Sent-BERT} \\ \cline{2-8} \cline{10-13} 
\multicolumn{1}{l|}{\multirow{-2}{*}{{\color[HTML]{333333} \textbf{Model}}}} & \multicolumn{1}{c|}{B1}                                 & \multicolumn{1}{c|}{B2}             & \multicolumn{1}{c|}{B3}             & \multicolumn{1}{c|}{B4}             & \multicolumn{1}{c|}{RL}             & \multicolumn{1}{c|}{R1}             & \multicolumn{1}{c|}{R2}             & \multicolumn{1}{c|}{\multirow{-2}{*}{\textbf{METEOR}}} & \multicolumn{1}{c|}{Pre}           & \multicolumn{1}{c|}{Rec}           & \multicolumn{1}{c|}{F1}            & (Cosine)           \\ \hline
\multicolumn{1}{l|}{w/o-Caption}                                                & \multicolumn{1}{c|}{22.85}                              & \multicolumn{1}{c|}{11.83}          & \multicolumn{1}{c|}{7.30}           & \multicolumn{1}{c|}{4.64}           & \multicolumn{1}{c|}{21.27}          & \multicolumn{1}{c|}{18.19}          & \multicolumn{1}{c|}{6.26}           & \multicolumn{1}{c|}{16.54}                             & \multicolumn{1}{c|}{86.40}          & \multicolumn{1}{c|}{86.10}          & \multicolumn{1}{c|}{86.20}          & 53.82              \\ 
\multicolumn{1}{l|}{w/-Visual}                                                  & \multicolumn{1}{c|}{49.97}                              & \multicolumn{1}{c|}{39.45}          & \multicolumn{1}{c|}{32.76}          & \multicolumn{1}{c|}{27.78}          & \multicolumn{1}{c|}{46.12}          & \multicolumn{1}{c|}{46.34}          & \multicolumn{1}{c|}{30.21}          & \multicolumn{1}{c|}{40.86}                             & \multicolumn{1}{c|}{90.10}          & \multicolumn{1}{c|}{89.70}          & \multicolumn{1}{c|}{89.90}          & 67.02              \\
\multicolumn{1}{l|}{w/o-Obj}                                                 & \multicolumn{1}{c|}{53.89}                              & \multicolumn{1}{c|}{43.18}          & \multicolumn{1}{c|}{36.65}          & \multicolumn{1}{c|}{31.86}          & \multicolumn{1}{c|}{49.13}          & \multicolumn{1}{c|}{50.48}          & \multicolumn{1}{c|}{34.53}          & \multicolumn{1}{c|}{49.38}                             & \multicolumn{1}{c|}{90.80}          & \multicolumn{1}{c|}{91.20}          & \multicolumn{1}{c|}{91.00}          & 72.27              \\ 

\multicolumn{1}{l|}{w/o-Graph}                                               & \multicolumn{1}{c|}{53.39}                              & \multicolumn{1}{c|}{42.90}          & \multicolumn{1}{c|}{36.08}          & \multicolumn{1}{c|}{31.65}          & \multicolumn{1}{c|}{48.17}          & \multicolumn{1}{c|}{50.25}          & \multicolumn{1}{c|}{34.21}          & \multicolumn{1}{c|}{49.21}                             & \multicolumn{1}{c|}{91.40}          & \multicolumn{1}{c|}{89.70}          & \multicolumn{1}{c|}{90.50}          & 71.77              \\

\multicolumn{1}{l|}{w/-FullGraph}                                                & \multicolumn{1}{c|}{32.84}                              & \multicolumn{1}{c|}{18.74}          & \multicolumn{1}{c|}{12.29}           & \multicolumn{1}{c|}{8.44}           & \multicolumn{1}{c|}{29.21}          & \multicolumn{1}{c|}{29.20}          & \multicolumn{1}{c|}{11.69}           & \multicolumn{1}{c|}{22.31}                             & \multicolumn{1}{c|}{87.10}          & \multicolumn{1}{c|}{87.30}          & \multicolumn{1}{c|}{87.40}          & 62.21             \\

\hline
\multicolumn{1}{l|}{\textbf{TEAM}}                                      & \multicolumn{1}{c|}{\textbf{55.32}}                     & \multicolumn{1}{c|}{\textbf{45.12}} & \multicolumn{1}{c|}{\textbf{38.27}} & \multicolumn{1}{c|}{\textbf{33.16}} & \multicolumn{1}{c|}{\textbf{50.58}} & \multicolumn{1}{c|}{\textbf{51.72}} & \multicolumn{1}{c|}{\textbf{34.96}} & \multicolumn{1}{c|}{\textbf{50.95}}                    & \multicolumn{1}{c|}{\textbf{91.80}} & \multicolumn{1}{c|}{\textbf{91.60}} & \multicolumn{1}{c|}{\textbf{91.70}} & \textbf{72.92}     \\ \hline

\multicolumn{13}{c}{(a) All samples}   \\ 

\hline
\multicolumn{1}{l|}{{\color[HTML]{333333} }}                                 & \multicolumn{4}{c|}{\textbf{BLEU}}                                                                                                                                        & \multicolumn{3}{c|}{\textbf{Rouge}}                                                                             & \multicolumn{1}{c|}{}                                  & \multicolumn{3}{c|}{\textbf{BERT-Score}}                                                                     & \textbf{Sent-BERT} \\ \cline{2-8} \cline{10-13} 
\multicolumn{1}{l|}{\multirow{-2}{*}{{\color[HTML]{333333} \textbf{Model}}}} & \multicolumn{1}{c|}{B1}                                 & \multicolumn{1}{c|}{B2}             & \multicolumn{1}{c|}{B3}             & \multicolumn{1}{c|}{B4}             & \multicolumn{1}{c|}{RL}             & \multicolumn{1}{c|}{R1}             & \multicolumn{1}{c|}{R2}             & \multicolumn{1}{c|}{\multirow{-2}{*}{\textbf{METEOR}}} & \multicolumn{1}{c|}{Pre}           & \multicolumn{1}{c|}{Rec}           & \multicolumn{1}{c|}{F1}            & (Cosine)           \\ \hline
\multicolumn{1}{l|}{w/o-Caption}                                                & \multicolumn{1}{c|}{23.31}                              & \multicolumn{1}{c|}{12.53}          & \multicolumn{1}{c|}{8.23}           & \multicolumn{1}{c|}{5.02}           & \multicolumn{1}{c|}{22.23}          & \multicolumn{1}{c|}{19.09}          & \multicolumn{1}{c|}{7.83}           & \multicolumn{1}{c|}{17.42}                             & \multicolumn{1}{c|}{87.50}          & \multicolumn{1}{c|}{87.30}          & \multicolumn{1}{c|}{87.40}          & 54.97              \\
\multicolumn{1}{l|}{w/-Visual}                                                  & \multicolumn{1}{c|}{50.29}                              & \multicolumn{1}{c|}{40.31}          & \multicolumn{1}{c|}{33.82}          & \multicolumn{1}{c|}{28.41}          & \multicolumn{1}{c|}{47.24}          & \multicolumn{1}{c|}{47.38}          & \multicolumn{1}{c|}{31.37}          & \multicolumn{1}{c|}{41.75}                             & \multicolumn{1}{c|}{90.50}          & \multicolumn{1}{c|}{90.10}          & \multicolumn{1}{c|}{90.30}          & 67.81              \\
\multicolumn{1}{l|}{w/o-Obj}                                                 & \multicolumn{1}{c|}{55.32}                              & \multicolumn{1}{c|}{44.87}          & \multicolumn{1}{c|}{37.82}          & \multicolumn{1}{c|}{33.96}          & \multicolumn{1}{c|}{50.58}          & \multicolumn{1}{c|}{52.45}          & \multicolumn{1}{c|}{36.12}          & \multicolumn{1}{c|}{51.06}                             & \multicolumn{1}{c|}{91.60}          & \multicolumn{1}{c|}{91.80}          & \multicolumn{1}{c|}{91.90}          & 72.98              \\

\multicolumn{1}{l|}{w/o-Graph}                                               & \multicolumn{1}{c|}{54.65}                              & \multicolumn{1}{c|}{43.82}          & \multicolumn{1}{c|}{37.29}          & \multicolumn{1}{c|}{32.27}          & \multicolumn{1}{c|}{50.42}          & \multicolumn{1}{c|}{51.18}          & \multicolumn{1}{c|}{35.26}          & \multicolumn{1}{c|}{49.25}                             & \multicolumn{1}{c|}{91.80}          & \multicolumn{1}{c|}{90.20}          & \multicolumn{1}{c|}{91.30}          & 72.31              \\

\multicolumn{1}{l|}{w/-FullGraph}                                                & \multicolumn{1}{c|}{33.56}                              & \multicolumn{1}{c|}{19.35}          & \multicolumn{1}{c|}{13.62}           & \multicolumn{1}{c|}{9.18}           & \multicolumn{1}{c|}{30.87}          & \multicolumn{1}{c|}{30.22}          & \multicolumn{1}{c|}{13.04}           & \multicolumn{1}{c|}{23.21}                             & \multicolumn{1}{c|}{87.20}          & \multicolumn{1}{c|}{87.40}          & \multicolumn{1}{c|}{87.50}          & 63.92             \\

\hline
\multicolumn{1}{l|}{\textbf{TEAM}}                                      & \multicolumn{1}{c|}{\textbf{56.45}}                     & \multicolumn{1}{c|}{\textbf{46.34}} & \multicolumn{1}{c|}{\textbf{39.58}} & \multicolumn{1}{c|}{\textbf{34.34}} & \multicolumn{1}{c|}{\textbf{52.79}} & \multicolumn{1}{c|}{\textbf{53.81}} & \multicolumn{1}{c|}{\textbf{36.78}} & \multicolumn{1}{c|}{\textbf{51.62}}                    & \multicolumn{1}{c|}{\textbf{92.40}} & \multicolumn{1}{c|}{\textbf{92.90}} & \multicolumn{1}{c|}{\textbf{92.30}} & \textbf{73.35}     \\ \hline
\multicolumn{13}{c}{(b) Non-OCR samples}                                                                                                                                                                                                                                                                                                                                                                                                                                                                                                                                \\ \hline
\multicolumn{1}{l|}{{\color[HTML]{333333} }}                                 & \multicolumn{4}{c|}{\textbf{BLEU}}                                                                                                                                        & \multicolumn{3}{c|}{\textbf{Rouge}}                                                                             & \multicolumn{1}{c|}{}                                  & \multicolumn{3}{c|}{\textbf{BERT-Score}}                                                                     & \textbf{Sent-BERT} \\ \cline{2-8} \cline{10-13} 
\multicolumn{1}{l|}{\multirow{-2}{*}{{\color[HTML]{333333} \textbf{Model}}}} & \multicolumn{1}{c|}{{\color[HTML]{333333} \textbf{B1}}} & \multicolumn{1}{c|}{B2}             & \multicolumn{1}{c|}{B3}             & \multicolumn{1}{c|}{B4}             & \multicolumn{1}{c|}{RL}             & \multicolumn{1}{c|}{R1}             & \multicolumn{1}{c|}{R2}             & \multicolumn{1}{c|}{\multirow{-2}{*}{\textbf{METEOR}}} & \multicolumn{1}{c|}{Pre}           & \multicolumn{1}{c|}{Rec}           & \multicolumn{1}{c|}{F1}            & (Cosine)           \\ \hline
\multicolumn{1}{l|}{w/o-Caption}                                                & \multicolumn{1}{c|}{21.35}                              & \multicolumn{1}{c|}{10.23}          & \multicolumn{1}{c|}{6.21}           & \multicolumn{1}{c|}{3.25}           & \multicolumn{1}{c|}{20.57}          & \multicolumn{1}{c|}{16.61}          & \multicolumn{1}{c|}{5.02}           & \multicolumn{1}{c|}{15.83}                             & \multicolumn{1}{c|}{85.20}          & \multicolumn{1}{c|}{85.10}          & \multicolumn{1}{c|}{85.40}          & 52.94              \\
\multicolumn{1}{l|}{w/-Visual}                                                  & \multicolumn{1}{c|}{48.37}                              & \multicolumn{1}{c|}{38.25}          & \multicolumn{1}{c|}{31.26}          & \multicolumn{1}{c|}{26.28}          & \multicolumn{1}{c|}{44.60}          & \multicolumn{1}{c|}{45.12}          & \multicolumn{1}{c|}{29.02}          & \multicolumn{1}{c|}{39.97}                             & \multicolumn{1}{c|}{89.90}          & \multicolumn{1}{c|}{89.50}          & \multicolumn{1}{c|}{89.70}          & 66.54              \\
\multicolumn{1}{l|}{w/o-Obj}                                                 & \multicolumn{1}{c|}{52.19}                              & \multicolumn{1}{c|}{42.86}          & \multicolumn{1}{c|}{35.24}          & \multicolumn{1}{c|}{31.02}          & \multicolumn{1}{c|}{46.88}          & \multicolumn{1}{c|}{49.60}          & \multicolumn{1}{c|}{33.17}         & \multicolumn{1}{c|}{48.46}                             & \multicolumn{1}{c|}{90.20}          & \multicolumn{1}{c|}{90.60}          & \multicolumn{1}{c|}{90.70}          & 71.64              \\

\multicolumn{1}{l|}{w/o-Graph}                                               & \multicolumn{1}{c|}{51.32}                              & \multicolumn{1}{c|}{41.91}          & \multicolumn{1}{c|}{34.25}          & \multicolumn{1}{c|}{31.23}          & \multicolumn{1}{c|}{46.57}          & \multicolumn{1}{c|}{49.26}          & \multicolumn{1}{c|}{33.97}          & \multicolumn{1}{c|}{49.18}                             & \multicolumn{1}{c|}{90.70}          & \multicolumn{1}{c|}{89.40}          & \multicolumn{1}{c|}{89.60}          & 70.31              \\

\multicolumn{1}{l|}{w/-FullGraph}                                                & \multicolumn{1}{c|}{32.13}                              & \multicolumn{1}{c|}{18.12}          & \multicolumn{1}{c|}{11.46}           & \multicolumn{1}{c|}{7.76}           & \multicolumn{1}{c|}{28.16}          & \multicolumn{1}{c|}{28.35}          & \multicolumn{1}{c|}{10.16}           & \multicolumn{1}{c|}{21.45}                             & \multicolumn{1}{c|}{86.80}          & \multicolumn{1}{c|}{87.10}          & \multicolumn{1}{c|}{87.30}          & 60.57             \\

\hline
\multicolumn{1}{l|}{\textbf{TEAM}}                                      & \multicolumn{1}{c|}{\textbf{52.88}}                     & \multicolumn{1}{c|}{\textbf{43.08}} & \multicolumn{1}{c|}{\textbf{36.81}} & \multicolumn{1}{c|}{\textbf{32.34}} & \multicolumn{1}{c|}{\textbf{48.46}} & \multicolumn{1}{c|}{\textbf{49.68}} & \multicolumn{1}{c|}{\textbf{33.83}} & \multicolumn{1}{c|}{\textbf{49.25}}                    & \multicolumn{1}{c|}{\textbf{90.90}} & \multicolumn{1}{c|}{90.00} & \multicolumn{1}{c|}{\textbf{90.80}} & \textbf{71.93}     \\ \hline
\multicolumn{13}{c}{(c) OCR samples}                        
\end{tabular}%
}
\end{table*}

\subsection{Dataset}
We conducted experiments on the multimodal sarcasm explanation dataset \textbf{MORE}~\cite{DBLP:conf/aaai/Desai0A22}. It is created by collecting sarcastic posts from various social media sites (Twitter\footnote{\url{https://twitter.com/home}.}, Instagram\footnote{\url{https://www.instagram.com/}.} and Tumblr\footnote{\url{https://www.tumblr.com/}.}), where the  sarcasm explanation for each post is manually annotated. 
Finally, this dataset contains $3,510$ triplets in the form of \textless $image, caption, explanation$\textgreater, including $2,983$ for training, $175$ for validation, and $352$ for testing. Statistics of this dataset are summarized in Table \ref{tab:table1}.

\subsection{Experimental Setup}
We adopted the bart-base-chinese model provided by huggingface\footnote{\url{https://huggingface.co/facebook/bart-base}.} as the backbone of our model.
In practice, the total number of tokens in each sample, \ie $N$, is unified to 256 by padding or truncation operations.
The feature dimension $D$ is set to $768$, and the largest number of objects we allow to extract from an image, \ie $K$, is set to $36$.
We used AdamW~\cite{DBLP:journals/corr/abs-1711-05101} as the optimizer and  set the learning rate of GCN layers to 1e-3 and that of the BART to 1e-4. 
The batch size is set to $16$ and the maximum number of epochs for model training is set to $20$.
Following the previous work~\cite{DBLP:conf/aaai/Desai0A22}, we employed \mbox{BLEU-1}, \mbox{BLEU-2}, \mbox{BLEU-3}, \mbox{BLEU-4}~\cite{DBLP:conf/acl/PapineniRWZ02}, \mbox{ROUGE-1}, \mbox{ROUGE-2}, \mbox{ROUGE-L}~\cite{lin-2004-rouge}, METEOR~\cite{DBLP:conf/acl/BanerjeeL05},  BERT-Score~\cite{DBLP:conf/iclr/ZhangKWWA20} and Sent-BERT~\cite{DBLP:conf/emnlp/ReimersG19} to evaluate the performance of text generation models.

\subsection{On Model Comparison}
To validate our TEAM, we compared it with the following existing methods.
\begin{itemize}
    \item \textbf{PGN}~\cite{DBLP:conf/acl/SeeLM17}.
    Pointer Generator Network is a text-based generation model, which generates the text with not only a conventional decoder but also a copy mechanism that copies words directly from input caption.
    \item \textbf{Transformer}~\cite{DBLP:conf/nips/VaswaniSPUJGKP17}.
    This is also a text-based generation baseline, which generates the text with the advanced transformer architecture.
    \item \textbf{MFFG-RNN} and \textbf{MFFG-Trans}. These are two variations of {MFFG}~\cite{DBLP:conf/emnlp/LiuSYZX20}, a multimodal-based generation model for video summarization, where 
    {MFFG-RNN} and {MFFG-Trans} adopt the RNN and transformer architecture as the decoder, respectively.
    \item \textbf{M-Transf}~\cite{DBLP:conf/acl/YaoW20}.
    To use the visual modality to improve the quality of multimodal machine translation, this model equips Transformer with the multimodal self-attention mechanism to avoid encoding irrelevant information in images.
    \item \textbf{ExMore}~\cite{DBLP:conf/aaai/Desai0A22}. This is the most relevant baseline, which is designed for the task of multimodal sarcasm explanation. This method adopts BART as the model backbone and employs the cross-modal attention to inject the visual information into BART. 
    \item  \textbf{TEAM-w/o-Know}. Considering that all the baselines do not use the external knowledge, for fair comparison, we also introduced this variant of our model, where all the knowledge concepts are removed from our model.
\end{itemize}

\begin{figure*}[h]
    \centering
    \includegraphics[width=\textwidth]{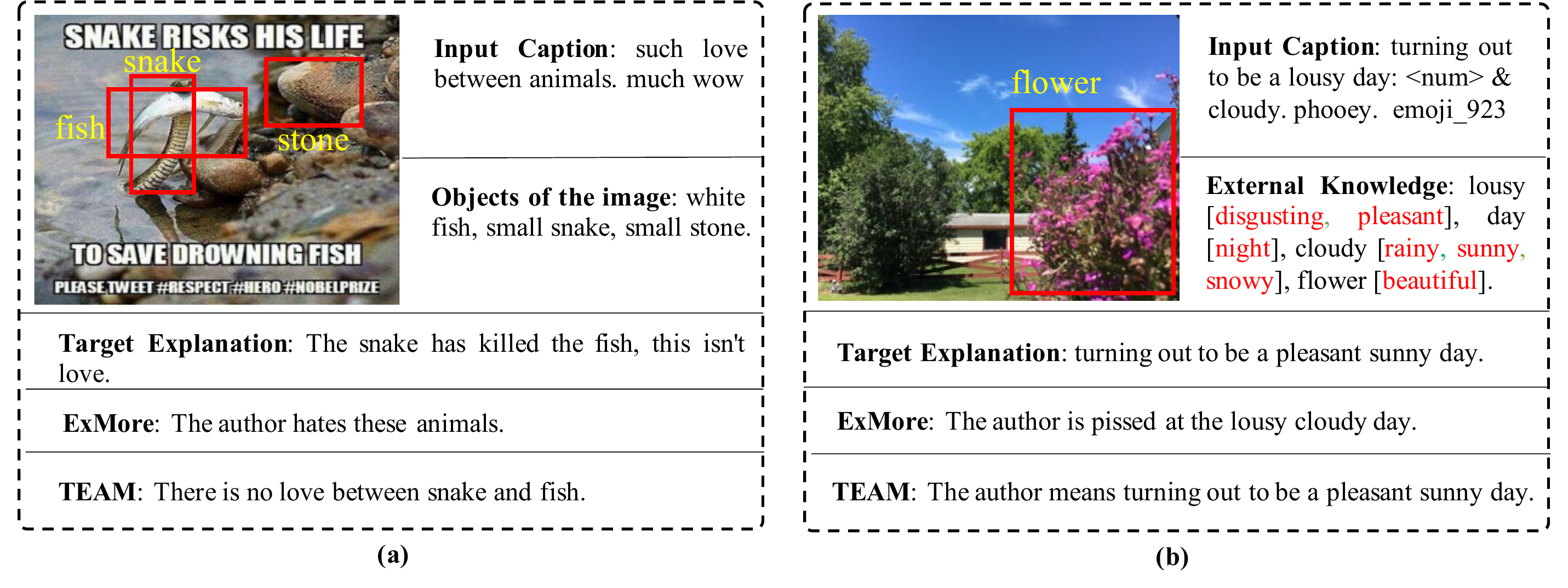}
     \vspace{-0.8cm}
    \caption{Comparison between the explanation generated by our model and the best baseline ExMore on two testing samples. The words in red are the related external knowledge concepts.}
    \label{fig:fig3}
\end{figure*}
Following the existing work~\cite{DBLP:conf/aaai/Desai0A22}, we conducted the performance comparison among different methods under three dataset configurations: a) on all samples, b) only on Non-OCR samples, and c) only on OCR samples.
OCR samples denote the samples whose images contain embedded texts, while Non-OCR samples do not. 
We reported the experiment results in Table~\ref{tab:table2}. From this table, we have several observations. 
(1) Both our complete model TEAM and its variant TEAM-w/o-Know consistently exceed all the state-of-the-art baselines in terms of all the metrics across different dataset configurations, which thoroughly demonstrates the superiority of our model. 
(2) The multimodal-based generation models (\eg  MFFG-RCNN and MFFG-Transf) do not always perform better than the text-based models (\eg PGN). This implies that the performance of the model could be worse if the visual modality is not used properly. 
3) The performance of our model on Non-OCR samples is higher than that on OCR samples across all metrics. The possible reason is that since our model only considers the object-level meta-data, the embedded text in the image could be ignored, leading to the information loss. In spite of this, our model still achieves a significant improvement over the best baseline on the Non-OCR samples.

\subsection{On Ablation Study}
We introduced the following variants of our model for the ablation study. 1) \textbf{w/o-Caption}. To evaluate the role of the caption in sarcasm explanation generation, we did not utilize the caption in this model. 2) \textbf{w/-Visual}. 
    To show the superiority of using the object meta-data over the object visual feature, we adopted the object visual features extracted by {Vit}~\cite{DBLP:conf/iclr/DosovitskiyB0WZ21}, and concatenated them with the textual caption features to derive $\mathbf{H}$, while the object meta-data is totally removed. 
     3) \textbf{w/o-Obj}. To show the benefit of extracting the key objects from the images, we omitted the object meta-data from the input.
4) \textbf{w/o-Graph}. To verify the necessity of building the multi-source semantic graph for sarcasm reasoning, we removed $\mathbf{G}_L$ and only fed $\mathbf{H}$ into the BART decoder.
  5) \textbf{w/-FullGraph}. To further investigate the semantic relations of our multi-source semantic graph, we erased all the semantic relations and transformed the semantic graph to a full connection graph.



The ablation study results are shown in Table~\ref{tab:ablation}. From this table, we have the following observations. 
1) w/o-Caption performs terribly compared with TEAM. This is reasonable since the caption is the main source for delivering the ironic intention. 
2) TEAM exceeds w/-Visual. It demonstrates that the object-level metadata is better than the visual feature to stimulate the generation of sarcasm explanation with BART. 
3) TEAM consistently outperforms w/o-Obj across different evaluation metrics. It confirms the necessity of using object-level feature for generating sarcasm explanation. 
4) TEAM outperforms w/o-Graph, denoting that the graphs are essential to capture the ironic intention in the multimodal sarcastic posts.
And 5) w/-FullGraph performs worse than TEAM, which verifies the utility of proposed semantic relations.
\subsection{On Case Study}

To get an intuitive understanding of how our model works on multi-modal sarcasm explanation, we showed two testing samples in Figure~\ref{fig:fig3} due to the limited space. For comparison, we also displayed the explanation results of the best baseline ExMore.
In case (a), as you can see, our model performs better than ExMore in terms of the quality of the generated sarcasm explanation. 
This may be attributed to the fact that our model considers the object-level metadata (\ie ``fish'' and ``snake'') of the image, which benefits the sarcasm reasoning and explanation generation.
In case (b), our model correctly explains the sarcasm, while ExMore failed.  
By analyzing the retrieved external knowledge concepts, we noticed that the concept ``disgusting'' benefits the semantic learning of the input caption, while concepts ``sunny'' and ``beautiful'' promotes the semantic interpretation of the input image. Moreover, the related concept ``pleasant'' of the word ``lousy'' contributes to the sarcasm explanation generation. Overall, these two cases intuitively show the benefits of incorporating both object-level meta-data and external knowledge concepts in the context of multimodal sarcasm explanation.

%% file: 5_conclusion.tex
\section{Conclusion and Future Work}
In this work, we propose a novel multi-source semantic graph-based multimodal sarcasm explanation generation scheme. 
Experimental results on a public dataset demonstrate the superiority of our model over existing cutting-edge methods, and validate the advantage of utilizing the object-level meta-data over the global visual feature of the image as well as the benefit of incorporating the external knowledge in the context of multimodal sarcasm explanation. 
Particularly, we notice that our model performs worse on OCR samples than on Non-OCR samples. This is due to that our model currently ignores the text embedded in the image. 
In the future, we plan to incorporate the embedded text, which could indicate important clues for sarcasm explanation, to boost the model performance.